\documentclass[runningheads]{llncs}
\usepackage{svg}
\usepackage{graphicx}
\usepackage[export]{adjustbox}
\usepackage{enumerate}
\usepackage{amssymb}
\usepackage[subrefformat=parens,labelformat=parens]{subfig}
\usepackage{amsmath}
\usepackage{multirow}
\usepackage{booktabs}
\usepackage{textgreek}
\usepackage{hhline}
\usepackage{array}
\usepackage{tabu}
\usepackage{verbatim} 
\usepackage{graphics}
\usepackage{color}
\usepackage{url}
\usepackage{ bbold }

\begin{document}
\title{Graph-based Deep Generative Modelling for Document Layout Generation}
\titlerunning{Graph-based Deep Generative Modelling for Document Layout Generation}
%
\author{Sanket Biswas\inst{1}\orcidID{0000-0001-6648-8270} \and
Pau Riba\inst{1}\orcidID{0000-0002-4710-0864} \and
Josep Lladós\inst{1}\orcidID{0000-0002-4533-4739} \and
Umapada Pal\inst{2}\orcidID{0000-0002-5426-2618}}
\authorrunning{S.Biswas et al.}
%
\institute{Computer Vision Center \& Computer Science Department  \\
              Universitat Autònoma de Barcelona, Spain \\
              \email{\{sbiswas, priba, josep\}@cvc.uab.es} 
              \and
              CVPR Unit, Indian Statistical Institute, India \\ 
              \email{umapada@isical.ac.in}}
\maketitle              
\begin{abstract}

One of the major prerequisites for any deep learning approach is the availability of large-scale training data. When dealing with scanned document images in real world scenarios, the principal information of its content is stored in the layout itself. 
In this work, we have proposed an automated deep generative model using Graph Neural Networks (GNNs) to generate synthetic data with highly variable and plausible document layouts that can be used to train document interpretation systems, in this case, specially in digital mailroom applications. It is also the first graph-based approach for document layout generation task experimented on administrative document images, in this case, invoices.

\end{abstract}
\section{Introduction}\label{s:intro}
 The variability and diversity of complex layouts and graphical entities in digital mailroom documents prevent us from tackling document understanding problems separately, and that such specificity has been a  great  barrier  towards  deriving off-the-shelf  document  analysis solutions, usable by nonspecialists. Apparently, OCR-based engines are the most widely recognized products in this research community. For instance, imagine a business firm having thousands of documents to process, analyze, and transform to carry out day-to-day operations. Examples of such documents might include receipts, invoices, forms, statements, contracts, and many more pieces of data, which are highly unstructured or semi-structured, and it is essential to be able to quickly analyze and understand the information embedded within the unevenly structured data in these cases. In most of these Document Image Analysis and Recognition(DIAR) applications, the document content has been broadly classified into two structural entities: (1) physical and (2) logical structural entities. While the physical structure describes the visual aspect of the document by representing the specific objects and their mutual positions, the logical structure assigns a definite semantic meaning to each of these objects.
 
In recent times, deep CNN-based methods have tried to deduct the visual differences between object classes: while the visual characteristics of certain graphical elements (e.g., plots, charts, figures) differ conspicuously from text elements, the same cannot be said for tables, where the major differences from the surrounding content lie mostly stored in the layout information and its context. Moreover, trying to train these deep CNN models from scratch may be quite impractical due to the requirement of a large amount of training examples and the need of precisely annotated document datasets, which are scarcely available in the community. The key reason may be that most of these documents (administrative documents, for example) contain sensitive information (identity name, bank details, health information and  so on) and are not publicly released by government agencies or business firms to be used in cloud services. Hence, as in many other applications requiring intensive training, data augmentation through synthetic generation is a solution. In the case of document structure recognition, there is an important need to generate synthetic document layouts that can encode the structural information of the real data and can be used during training to transfer enough knowledge to the model. Patil et. al. \cite{gadi2020read} formulated this task as Document Layout Generation (DLG), where they used a recursive neural network approach to map the structured representation of semi-structured documents  (in the form of tree-level hierarchies) to a code representation, the space of which is approximated by a Gaussian. New hierarchies representing plausible 2D document layouts were sampled from such distributions. In this work, we tackle the problem by encoding the structured hierarchies in the form of graph representation. 

Graphs possess the ability to represent two types of contextual knowledge: (1) geometric/intrinsic, spatial structure of the document with positional information of object categories like tables, and (2) semantic/extrinsic, conceptual connections between the different object categories in a document. Therefore, graphs emerge as a suitable model to represent document layouts. The revolution of deep learning has also seen considerable progress in the area of graph-based representation and learning. Graph Neural Networks (GNNs) \cite{kipf2016semi,defferrard2016convolutional} as deep learning approaches have extended the power of CNNs to non-Euclidean geometries to capture long distance / different levels semantics based on the relations between objects. GNN's eventually learn a state embedding that contains the neighborhood information for each node (which can represent different entities or objects). The embedding is constructed at different graph convolution layers, so it encodes the information of a subgraph centered at a node. In the scenario of document interpretation, GNNs embed a description of a local layout as a context of a given document element. A recent application example was the detection of tables in case of administrative document images~\cite{riba2019table,qasim2019rethinking}.

\begin{figure}[h]
    \centering
    \includegraphics[width=\linewidth]{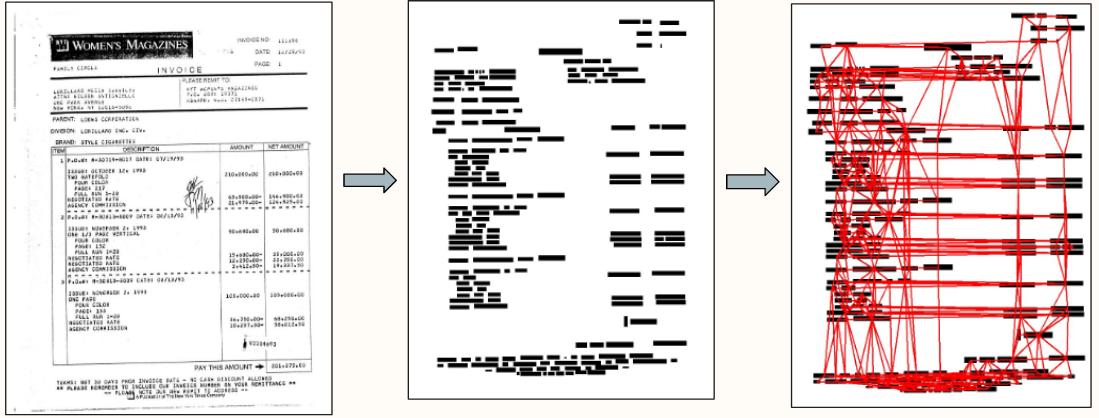}
    \caption{Graph representation of the structure of an invoice image}
    \label{fig:graph_rep}
\end{figure}

In summary, the main contributions of this work are as follows: (1) a  novel approach has been proposed for DLG task using GNN's to generate synthetic data applied to administrative invoice documents where  we  render  data  in  the  form  of  diverse  graphs  that  can  actually match  the structural characteristics of the target data. (2) The proposed graph-based generative modelling for such administrative documents also helps to invoke anonymity for the sensitive information (e.g. names, addresses, billing information, total amount etc.) it might contain in the document images. As shown in Fig. \ref{fig:graph_rep}, the nodes in the graph represent the different entities (e.g. header, table, supplier etc.) in the document, while the edges represent the visibility relations (horizontal or vertical) between the neighbouring nodes. (3) All experiments of our model have been performed on administrative invoices collection from the RVL-CDIP \cite{harley2015evaluation} dataset. As a result, a new synthetic invoice dataset has been created for augmenting the train data during table detection and layout analysis tasks. 

The rest of this paper is organized as follows. Section \ref{s:soa} provides a review of the relevant state of the art. In section \ref{s:method} we describe the main contribution of our work. Section \ref{s:results} provides experimental validation with some relevant results of our proposed approach, both qualitatively and quantitatively. Finally, Section \ref{s:conclusion} concludes the work throwing some light on its future scope and benefits.

\section{Related Work}\label{s:soa}

\subsection{Geometric Deep Learning}

Geometric deep learning~\cite{defferrard2016convolutional,kipf2016semi} has emerged 
as an extension of deep learning models to non-Euclidean domains, such as graphs and manifolds. To refer to neural networks applied to graph-structured data, the term Graph Neural Networks~\cite{scarselli2008graph} was coined.

The GNN methods help to learn representations at the node, edge and graph level considering the underlying topological information. Based on the fundamental architecture, GNN methods
can be aptly divided into two categories: spatial and spectral methods. Spatial methods extend the idea of Convolutional Neural Networks (CNNs) for images and define a set of operations involving the local neighbourhood to compute a new representation \cite{duvenaud2015convolutional,li2015gated}. On the other hand, spectral methods use the knowledge of spectral graph theory \cite{shuman2013emerging} and consider graph Laplacians for defining convolution operations in graph domain \cite{defferrard2016convolutional,kipf2016semi}. Gilmer et al. \cite{gilmer2017neural}
generalized both the domains of GNN, and defined their approach in terms of a Neural Message Passing (NMP) pipeline. These fundamental architectures have been further extended to new tasks involving graphs, such as the generative variational graph autoencoder \cite{kipf2016variational}, learning graph edit distance between a pair of graphs \cite{riba2018learning}, graph matching \cite{zanfir2018deep}, etc.

\subsection{Document Layout Generation}

The study and analysis of the structural properties and relations between entities in documents is a fundamental challenge in the field of information retrieval. Although local tasks like the Optical
Character Recognition (OCR) have been addressed with a considerably high model performance, the global and highly variable nature of document layouts has made their analysis some what more ambiguous. Previous works on structural document analysis mostly relied on the different kinds of specifically devised methods and applications~\cite{baird2012structured,breuel2003high,kasturi2002document,o1993document}. Recent works have shown that
deep learning based approaches have significantly improved the performance of these models in quality. A very standard approach in this regard was proposed by Yang et al. \cite{yang2017learning} which uses a joint visual and textual representation in a multimodal way of understanding, viewing the layout analysis as a pixel-wise segmentation task. But such modern deep learning based approaches typically require a very heavy amount of high-quality training data, that often calls for suitable methods to synthetically generate documents with real-looking layout \cite{li2019layoutgan} and content \cite{liu2018learning}. Our work actually focuses on the direction of research on synthetic layout generation, showing that our generated synthetic data can be extremely beneficial to augment training data for document analysis tasks. 

Preserving the reliable representation of layouts has shown to be very useful in various graphical design contexts, which typically involve highly structured and content-rich objects. One such recent intuitive understanding was established by Li et al. \cite{li2019layoutgan} in their LayoutGAN, which aims to generate realistic document layouts using Generative Adversarial Networks (GANs) with a wireframe rendering layer. Zheng et. al. \cite{zheng2019content} used a GAN-based approach to generate document layouts but their work focused mainly on content aware generation, that primarily uses the content of the document as an additional prior. Biswas et. al. \cite{biswas2021docsynth} devised a  generative GAN-based model to synthesize realistic document images, guided by a spatial layout(bounding boxes with object categories) given as a reference by the user.
However to use a more highly structured object generation, it is very important to focus operate on the low dimensional vectors unlike CNN's. Hence, in the most recent literature, Patil et. al. \cite{gadi2020read} has exploited this highly structured positional information along with content to generate document layouts. They have used recursive neural networks which operate on the low dimensional vectors and employ two-layer perceptrons to merge any two vectors, which make them computationally cheaper and help them train with fewer samples. The recursive neural networks are coupled with Variational Autoencoders (VAEs) in their resulting model architecture and provides state-of-the-art results for generating synthetic layouts for 2D documents. They have also introduced a novel metric for measuring document similarity, called DocSim, and used this metric to show the novelty and diversity of the generated layouts. 

Using geometric relations between the different entities in documents can actually help to preserve the structural information along with the content as seen in the work by Riba et. al.\cite{riba2019table} on table detection in invoice documents using GNNs. Figure \ref{fig:graph_rep} clearly illustrates how they have used graph modelling for document images to capture the geometrical structure of an invoice and using this knowledge can help us to generate more realistic synthetic samples for training. In this work we have used a similar kind of graph modelling for exploiting the structural information of an invoice image. Carbonell et. al.\cite{carbonell2021named} also used GNNs for recognition of structural components like named entities in semi-structured administrative documents.  Traditional generative models for graphs \cite{white2007spectral} are usually hand‐crafted to model a particular family of graphs, and thus they do not have the capacity to directly learn the generative model from observed data. To find a solution, one such graph-based generative model using GNNs was proposed by You et. al. \cite{you2018graphrnn} on molecular data generation. They used sequential generation with Recurrent Neural Networks (RNNs) on top of graph based representations and get state-of-the-art results on molecular data generation. But there has not been any substantial work in the literature which has applied such graph-based generative models for document layout analysis tasks. In this context, it is indeed a challenging problem which we tackle in this work. As case study, we will work in the context of administrative  documents, primarily focusing on invoices. Automated generation of synthetic document layouts will allow us to train document interpretation systems in a more efficient way for all kinds of document layout analysis tasks.

\section{Method}\label{s:method}
\begin{figure}[ht]
    \centering
    \includegraphics[width=\linewidth]{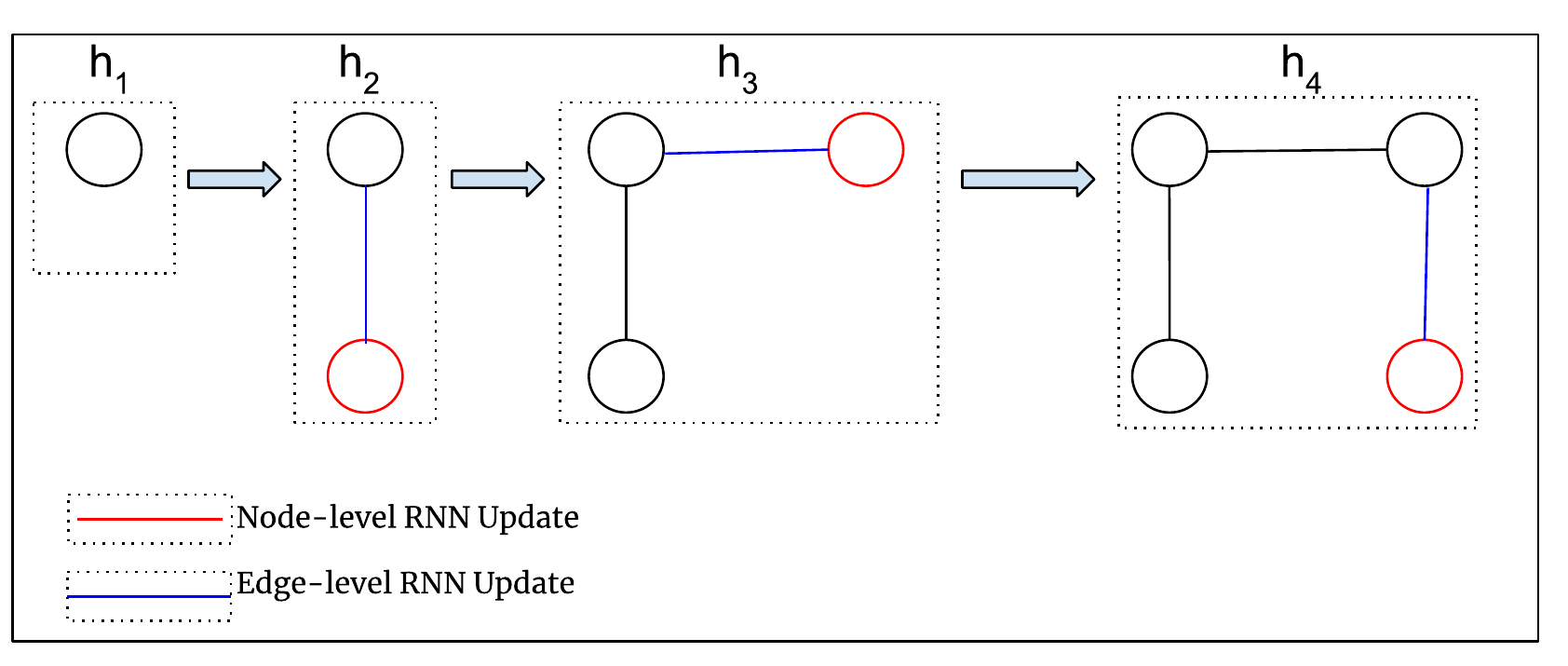}
    \caption{\textbf{The Graph Generation Process:} Given figure illustrates the graph generation process during inference time. The Node-level RNN update encodes the graph hidden state $h$ , updated by the predicted adjacency vector $S_{i}^{\pi}$ for every node. The Edge-level RNN updates the sequence of edges when every new node is added.} 
    \label{fig:graph_generation_process}
\end{figure}

\begin{figure}[ht]
    \centering
    \includegraphics[width=\linewidth]{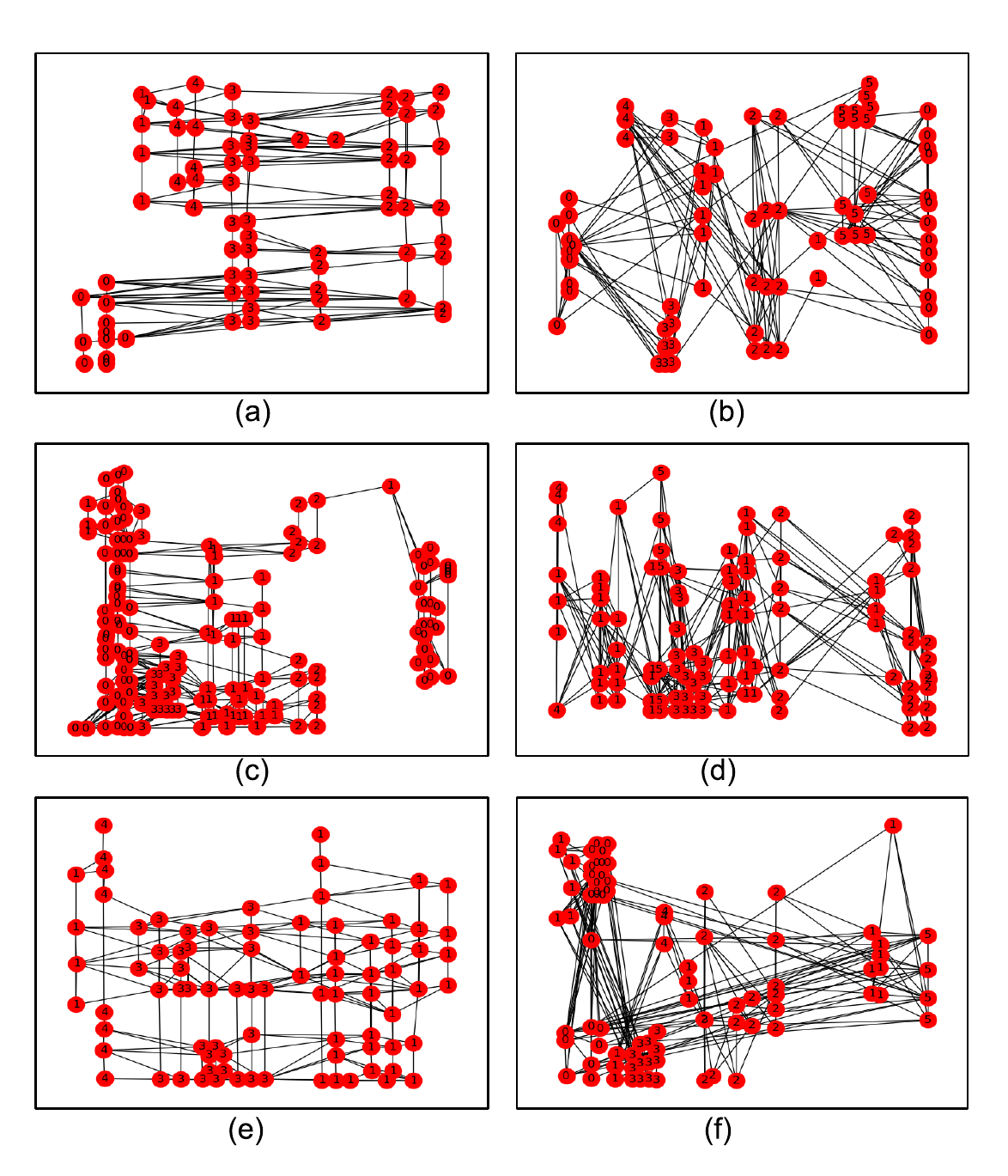}
    \caption{Qualitative analysis of our generative model with figures (a), (c) and (e) representing the test graphs from real invoice data and (b), (d) and (f) representing the generated graphs from our model.}
    \label{fig:qualitative_analysis}
\end{figure}

In this work, we have explored a new research direction in the DIAR domain using the application of GNN. A hierarchical and scalable framework has been designed and implemented to exploit highly powerful graph representations in semi-structured administrative documents like invoices. Every document image has been modelled as a visibility graph which is fed to our generative model to synthesize meaningful document layouts. Figure \ref{fig:graph_rep} depicts the structure of an invoice document and how it has been modelled as a visibility graph.
We considered a document graph whose nodes are graphical or named entities such as tables, figures, header, date, etc. while the edges represent their spatial relationships. We aim to learn a distribution $p(G)$ over an undirected graph $G=(V,E)$ defined by the node set $V=\left\{v_{1}, \ldots, v_{n}\right\}$ and edge set $E=\left\{\left(v_{i}, v_{j}\right) \mid v_{i}, v_{j} \in V\right\}$ that has a node ordering $\pi$ to map nodes to rows/columns of adjacency matrix $A^{\pi}$. This node ordering scheme has been adapted to enhance time efficiency of our training model. An adjacency matrix $A_{i, j}^{\pi}$ under a node ordering $\pi$ can be represented as $A_{i, j}^{\pi}=\mathbb{1}\left[\left(\pi\left(v_{i}\right), \pi\left(v_{j}\right)\right) \in E\right]$. 
So,in this work, we have proposed a graph generation framework applied in context to document images, that learns to generate realistic graphs by training on a representative set of graphs known as visibility graphs modelled from documents.

 The main idea is to represent graphs of different node orderings as sequences, and then build a generative model on top of these sequences. As illustrated in the graph generation framework in Fig.
 \ref{fig:graph_generation_process}, we decomposed the entire process into two parts: one that generates a sequence of nodes (Node-level update) and then another process that generates a sequence of edges for every new generated node (Edge-level update) which will be explained in more detail in the below subsections in a step-wise manner. 

\subsection{Graph to Sequence Mapping}
We aim to learn a distribution $p_{model}(G)$ over graphs , based on a set of observed graphs $G={(G_1,...,G_s)}$, sampled from a data distribution $p(G)$, where each graph, may have a different set of nodes and edges. 
During training time instead of learning $p(G)$ directly, whose sample space is really complex to define, we instead sample a node ordering $\pi$ to get a set of sequences $S^{\pi}$ as our observations and learn $p(S^{\pi})$ instead. This help us to learn a model autoregressively due to the $S^{\pi}$ . At inference time, we can simply sample G directly by computing $p(G)$ without this mapping. 
The mapping function $f_S$ from graphs to sequences, for a graph $G \sim p(G)$ with $n$ nodes under node ordering $\pi$ can be determined in equation \ref{sequence_mapping}.

\begin{equation}
\begin{split}
S^{\pi}=f_{S}(G, \pi)=\left(S_{1}^{\pi}, \ldots, S_{n}^{\pi}\right)
\end{split}
\label{sequence_mapping}
\end{equation}
Every element $S_{i}^{\pi}$ is actually an adjacency vector that represents the edges between the present node $\pi(v_i)$ and previous nodes $\pi(v_j)$ already in graph. This can be further represented by the equation \ref{sequence_adjacency}.
\begin{equation}
\begin{split}
S_{i}^{\pi}=\left(A_{1, i}^{\pi}, \ldots, A_{i-1, i}^{\pi}\right)^{T}, \forall i \in\{2,3,4 \ldots, n\}\end{split}
\label{sequence_adjacency}
\end{equation}
\subsection{GRNN Framework}

As a result of the graph to sequence mapping, our next step would be to generate the adjacency matrix of a graph G by generating these adjacency vectors $A_{i, i}^{\pi}$ of each node in a step by step sequential process. This can output different networks with variable number of nodes, while preserving the important topological properties of the generated graph. While transforming the learning distribution from $p(G)$ to $p(S^{\pi})$, we can further decompose $p(S^{\pi})$ as the product of conditional distributions over the elements due to its sequential nature as shown in equation \ref{learning_distribution}. 

\begin{equation}
\begin{split}
p\left(S^{\pi}\right)=\prod_{i=1}^{n+1} p\left(S_{i}^{\pi} \mid S_{1}^{\pi}, \ldots, S_{i-1}^{\pi}\right)\end{split}
\label{learning_distribution}
\end{equation}

Now to model this still complex distribution, we used recurrent neural networks (RNNs) that consist of a state-transition function and an output-function as shown in equations \ref{rnn_state} and \ref{rnn_output}: 

\begin{equation}
\begin{split}
h_{i}=f_{\text {trans }}\left(h_{i-1}, S_{i-1}^{\pi}\right)\end{split}
\label{rnn_state}
\end{equation}

\begin{equation}
\begin{split}
\theta_{i}=f_{\text {out }}\left(h_{i}\right)\end{split}
\label{rnn_output}
\end{equation}
where $h_{i}$ is a vector that encodes the updated generated graph information, $S_{i-1}^{\pi}$ is the adjacency vector for the last updated node $i-1$ and $\theta_{i}$ denotes the distribution of next node's adjacency vector i.e. $S_{i}^{\pi} \sim {P}_{\theta_{i}}$. 
So theoretically, the proposed Graph Recurrent Neural Network (GRNN) framework utilizes a hierarchical RNN as shown in figure \ref{fig:graph_generation_process}, where the first (i.e. graph-level) RNN generates nodes and update the state of the graph. The second RNN (i.e. edge-level) generates the edges of a given node. To achieve a scalable modeling, we let these networks share their weights across all the time steps $i$ during the training phase. 

\subsection{Learning via Breadth-First Search} 
A great insight in our proposed method is rather than learning to generate graphs with using the Breadth-First Search (BFS) node orderings, instead of random node permutations. The BFS function takes a random permutation $\pi$ as input, picks $\pi(v_1)$ as starting node and appends the node neighbours into the BFS queue in an order defined by  $\pi$. The modified equation for mapping function from graphs to sequences can rewritten as shown in equation \ref{bfs_output}. 
\begin{equation}
    S^{\pi}=f_{S}(G, \operatorname{BFS}(G, \pi))
\label{bfs_output}
\end{equation}

This technique helps the model to be trained on all possible BFS orderings, instead of all possible node permutations. As the BFS function is deterministic in nature and many-to-one, i.e. one same ordering can eventually map multiple permutations, it reduces number of sequences we need to consider. It also makes the learning much simpler by reducing the number of edge predictions and adding possible edges only for the nodes that are considered in the BFS queue itself.

\section{Experimental Validation}\label{s:results}


For our experimental validation, we have used the RVL-CDIP dataset\cite{harley2015evaluation}. In particular, the invoice subset has been split into 70 and 30 percent of the samples for training and evaluation respectively. Additionally, two extra datasets have been evaluated, namely, Protein \cite{borgwardt2005protein} and Community\cite{kim2015community}, to evaluate the robustness of our method on other domains.

\subsection{Datasets}

\subsubsection{RVL-CDIP Invoices~\cite{harley2015evaluation}.} 
The RVL-CDIP (Ryerson Vision Lab Complex Document Information Processing) is a well-known document information database containing 16 classes of documents with about 400,000 images in grayscale. For evaluating our synthetic graph generation framework, we chose the 518 images from the Invoice class, annotated with 5 different regions belonging to class header, table, address and so on. Here each invoice page is represented by a graph, with the nodes corresponding to the entities of different class in the invoice.    

\subsubsection{Protein~\cite{borgwardt2005protein}.}
918 protein graphs have been used with every node representing an amino acid and two nodes are connected if they are at a distance threshold of 6 Angstrom. 

\subsubsection{Community~\cite{kim2015community}.}
The community dataset contains a collection of 500 two-community graphs, where each community generated by Erdos-Renyi model (E-R) \cite{erdHos1960evolution}, represents a node in the graph. 
\subsection{Data Preparation for Graph Representation} In this stage, given the image of a document, we apply physical layout techniques to detect the graphical regions. Given an invoice document, we represent each detected entity corresponding to a 7-dimensional vector containing the position of the bounding boxes and the histogram of its content (numbers, alphabets or symbol). This encoded information will be used to generate a visibility graph in order to represent the structural information of the document. 
We consider $G=(V,E)$ to be a visibility graph. The set of edges $E$ represent visibility relations between nodes. Two entities are said to be connected with an edge if and only if the bounding boxes are vertically or horizontally visible, i.e. a straight horizontal or vertical line can be traced between the bounding box of two entities without crossing any other. Also, long edges covering more than a quarter of the page height are discarded. An example of a visibility graph sample with corresponding node embedding is shown in Figure \ref{fig:graph_rep}. 

\subsection{Training setup for Graph Recurrent Neural Network} Once the document has been processed and visibility graph has been generated, we feed them to our Graph Recurrent Neural Network (GRNN) model framework with the 7-dimensional node input space to get projected to a higher order space encoding with individual node features preserving the structural content information of the document. The graph-level RNN used in our work uses 4 layered GRU with 128 dimensional hidden state. To output the adjacency vector prediction, the edge-level RNN uses 4 layered GRU cells with 16 hidden dimensional state. To get the predicted adjacency vector in the output, the edge-level RNN maps the 16 dimensional hidden state to a 8 dimensional vector through a MLP and ReLU activation, then another MLP maps the vector to a scalar with sigmoid activation. We initialize the the edge-level RNN by the output of the graph-level RNN when generating the start of sequences $S_{i-1}^{\pi}$. We use the highest layer hidden state of the graph-level RNN to initialize with a linear layer to match the dimensionality. During the training time, ground truth has been used rather than the model's own predictions. During the inference time, the model is allowed to use its own predicted graph samples at each time step to generate a graph. The Adam Optimizer has been used for minibatch size of 32. We set the learning rate to be 0.001 which is decayed by 0.2 at every 100th epoch in all experiments. 

\subsection{Evaluation Schema}
The evaluation of the quality of generated graphs is quite hard to estimate. A fair comparison between the test graph and generated graph is required. By visualizing the sets of test graphs and the generated graphs, a fair qualitative comparison can be done. From Figure \ref{fig:qualitative_analysis} we can infer a qualitative comparison between the test and generated samples. 

For a quantitative evaluation scheme, we have used the Maximum Mean Discrepancy (MMD) measures to calculate the distance between the two sets of graphs (in this case, the test sample and the generated sample). In our experiments, the derived MMD scores between the graphs have been calculated for degree and clustering coefficient distributions, along with the average orbit count statistics as shown in Table \ref{tab:graph_gen}. The lower the scores, the better the real structure of the entities has been preserved. 

\subsection{Experiments on Administrative Invoice Documents} 

Experiments on the subset of administrative document (invoices) taken from RVL-CDIP\cite{harley2015evaluation} has been conducted and we report the first baseline for document layout generation using a Graph Neural Network(GNN) framework. 

As illustrated in Figure \ref{fig:qualitative_analysis}, we illustrate some of the qualitative results with the document graphs that we generated from our proposed GRNN model. The visualizations of the graph samples suggest that the generated graphs visually preserve the appearance of the reference one, so the model roughly learns to preserve both syntactic and semantic information for different entities. Eventually, we can create synthetic samples of invoices by generating more and more graph samples and also providing them during the inference time. Since this is the first baseline approach to use graph generative models in document datasets.

However, Table \ref{tab:graph_gen} depicts the quantitative results we obtained for RVL-CDIP Invoice dataset and we compared our model performance with some molecular datasets like Protein~\cite{borgwardt2005protein} and Community~\cite{kim2015community} present in the graph literature. Results clearly show that there is a huge room for improvement in the graph generative framework for documents when compared to the performance in the above mentioned benchmark molecular datasets. The `tables' entity is a regular structured entity and our model works well for generating table classes in realistic positions. But the title, date and other entities in administrative documents do not contain uniform information about its structural relations and its quite difficult for the model to learn those semantic content.

\begin{table}[ht]
\centering
\caption{Summary of the final model results for Document Layout Generation}
\label{tab:graph_gen}
\setlength{\tabcolsep}{8pt}
\begin{tabular}{lccc}
\toprule
Dataset           & Degree ($\downarrow$)  & Clustering ($\downarrow$) & Orbit ($\downarrow$) \\ \midrule
Protein~\cite{borgwardt2005protein}           & 0.014 & 0.002 & 0.039 \\
Community~\cite{kim2015community}         & 0.034 & 0.102 & 0.037 \\
RVL-CDIP Invoices~\cite{harley2015evaluation} & 0.373 & 0.166 & 0.188 \\ \bottomrule
\end{tabular}

\end{table}

\section{Conclusion}\label{s:conclusion}
In this work, we have presented a novel approach to automatically synthesize document layouts structures. The proposed method  is able to understand the complex interactions among the different layout components and generate plausible layouts for 2D documents. The graph-based generative approach also explores the power of GNN's towards the learning and generation of complex structured layouts for administrative invoices as a case study. 

The future scope of this work will be mainly focused on two research  lines. Firstly, there is a requirement for a more efficient evaluation of synthetically generated layouts when compared to real document layout samples both quantitatively and qualitatively. Secondly, exploiting this generated layout samples for supervision purposes can enhance the performance on well-defined tasks such as table detection or document layout analysis.

\section*{Acknowledgment}

This work has been partially supported by the Spanish projects RTI2018-095645-B-C21, and FCT-19-15244, and the Catalan projects 2017-SGR-1783, the CERCA Program / Generalitat de Catalunya and PhD Scholarship from AGAUR (2021FIB-10010). We are also indebted to Dr. Joan Mas Romeu for all the help and assistance provided during the data preparation stage for the experiments. 

\bibliographystyle{splncs04}
\bibliography{references}
%




\end{document}